\documentclass[conference]{IEEEtran}
\IEEEoverridecommandlockouts

\usepackage{cite}
\usepackage{amsmath,amssymb,amsfonts}
\usepackage{algorithmic}
\usepackage{graphicx}
\usepackage{textcomp}
\usepackage{xcolor}
\def\BibTeX{{\rm B\kern-.05em{\sc i\kern-.025em b}\kern-.08em
T\kern-.1667em\lower.7ex\hbox{E}\kern-.125emX}}
\usepackage{hyperref}
\usepackage{multirow}
\usepackage{subfigure} 

\begin{document}
\def\algorithmname{FGU3R}

\title{\algorithmname: Fine-Grained Fusion via Unified 3D Representation for Multimodal 3D Object Detection}

\author{
    \IEEEauthorblockN{
    Guoxin Zhang$^1$ \qquad
    Ziying Song$^2$ \qquad
    Lin Liu$^2$ \qquad
    Zhonghong Ou$^1$\IEEEauthorrefmark{1}
    }
    $^1$
    State Key Laboratory of Networking and Switching Technology, \\ Beijing University of Posts and Telecommunications \\
    $^2$
    School of Computer Science and Technology, Beijing Jiaotong University\\
    \thanks{
    \IEEEauthorrefmark{1} indicates the corresponding author (zhonghong.ou@bupt.edu.cn).
    
    This work is supported by National Natural Science of China under the Grant 62076035.}
}

\maketitle

\begin{abstract}
Multimodal 3D object detection has garnered considerable interest in autonomous driving. 
However, multimodal detectors suffer from dimension mismatches that derive from fusing 3D points with 2D pixels coarsely, which leads to sub-optimal fusion performance. 
In this paper, we propose a multimodal framework~\algorithmname~to tackle the issue mentioned above via unified 3D representation and fine-grained fusion, which consists of two important components. 
First, we propose an efficient feature extractor for raw and pseudo points, termed Pseudo-Raw Convolution (PRConv), which modulates multimodal features synchronously and aggregates the features from different types of points on key points based on multimodal interaction. 
Second, a Cross-Attention Adaptive Fusion (CAAF) is designed to fuse homogeneous 3D RoI (Region of Interest) features adaptively via a cross-attention variant in a fine-grained manner. 
Together they make fine-grained fusion on unified 3D representation. 
The experiments conducted on the KITTI and nuScenes show the effectiveness of our proposed method. 
\end{abstract}

\begin{IEEEkeywords}
3D object detection, multimodal, cross attention
\end{IEEEkeywords}

\section{Introduction}
3D object detection, which aims to intelligently predict the categories, locations, and sizes of objects in 3D space, plays a crucial role in many fields, e.g., robotics vision and autonomous driving~\cite{Temporal_object_track, liu2023caltracker, hybrid_c_t}. In past years, although LiDAR-only detectors~\cite{pointrcnn,pvrcnn,voxelrcnn,dtssd,imfusion} made huge achievements, the performance is sub-optimal due to the inherent flaws of LiDAR point clouds, e.g., sparsity and semantic poverty.
Recent 3D detectors try to introduce multimodal to overcome the issues of LiDAR-only detectors. 
\cite{frustum-pointnets, frustum-pointpillars, frustum-convnet} utilize mature 2D detectors to accurate frustums in 3D space to erase redundant background points. However, their performances are bounded by the 2D detector. 
\cite{MV3D,AVOD,contfuse,SCANet,graphalign, isfusion,graphbev} aim to transform point clouds into 2D representations for dimensional matching with 2D images via viewpoint transforming, which allows 2D convolution to extract features. 
PointPainting\cite{pointpainting} appends semantic scores, generated by semantic segmentation, to the corresponding raw point via the sensor calibration projection. 
\cite{mvx-net,epnet,pi-rcnn} establish the relationship between the features of point clouds and RGB images to explore fine-grained fusion.
\cite{deepfusion,cat-det} introduces the attention mechanism to fuse point and RGB pixel at the feature level. MMF~\cite{mmf} exploits multiple related tasks to complement features for the 3D task. 
FUTR3D~\cite{futr3d} exploit transformer to encode the features of modality-agnostic implicitly. 
\begin{figure}[!t]
    \centering 
    \subfigbottomskip=2pt
    \subfigcapskip=-3pt
    \subfigure[3D + 2D]{
        \includegraphics[width=0.98\linewidth]{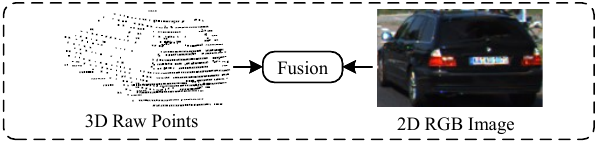}
    } \\
    \subfigure[3D + 3D (\textbf{Ours})]{
        \includegraphics[width=0.98\linewidth]{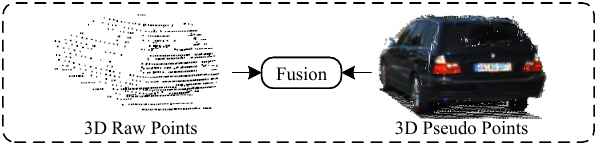}
    }
    \caption{
    (a) Due to the discrepancy between 3D points and 2D images, the feature of dimension mismatch is hard to fusion and align efficiently, resulting in sub-optimal integration performance.
    (b) The unified 3D representation we employ can fine-grained fuse easily while maintaining semantic adjacency.
    }
\label{fig:dem_gap}
\end{figure}

Although previous methods achieve impressive improvement, they suffer from two key problems. First, many methods~\cite{mvx-net,focalsconv,epnet,centerfusion} combine 3D points and 2D pixels---features which are not from the same dimension—--thereby posing dimension mismatch problems, as shown in Fig.~\ref{fig:dem_gap}a. 
Second, previous methods~\cite{MV3D,mvx-net,focalsconv,epnet,centerfusion} fuse simply multimodal features via elements-wise add or concatenation, which is a coarse-grained fusion that degrades detection performance.

\begin{figure*}[!t]
    \centering
        \includegraphics[width=\linewidth]{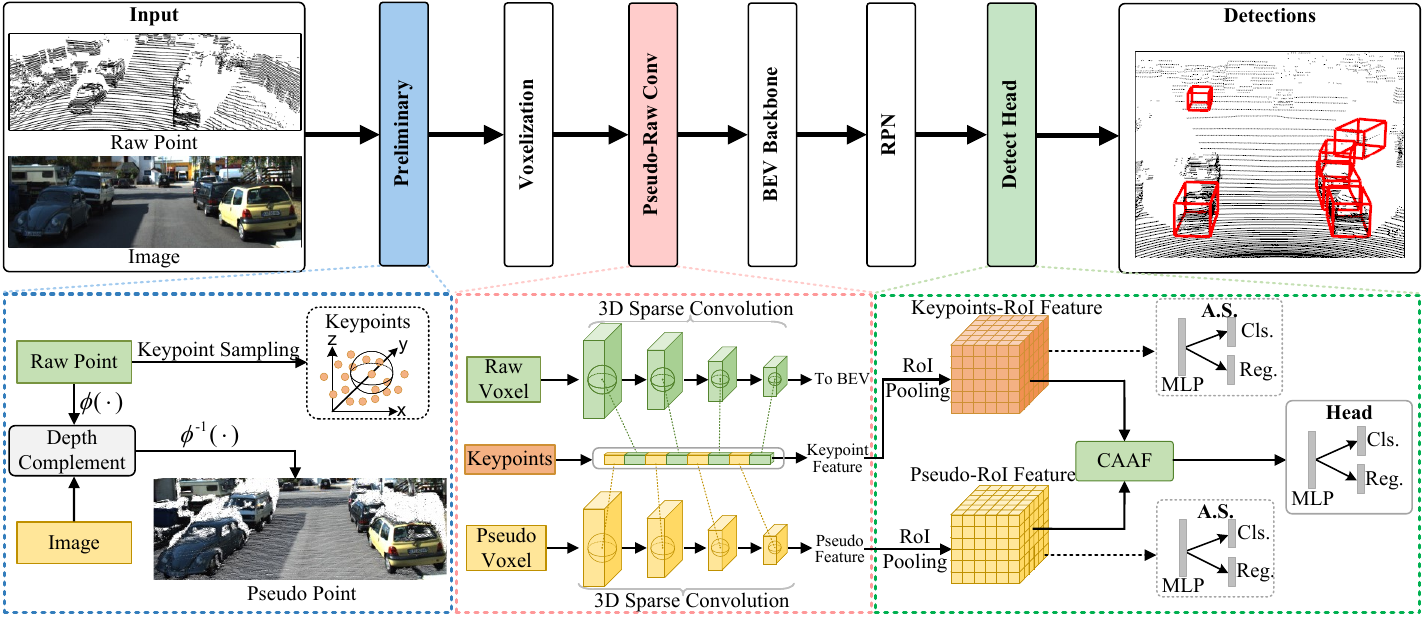}
        \caption{The overall architecture of \algorithmname.The dashed line means inference-only. RPN, BEV, and A.S. represent Region Proposal Network, Brid-eye's view, and Auxiliary Supervise. }
    \label{fig:framework}
\end{figure*}

To tackle the above challenge, we propose a multimodal 3D detection framework that utilizes explicit unified 3D representation to perform fine-grained fusion. 
\textbf{First}, we overcome dimension mismatch by providing explicit unified 3D representation, as shown in Fig.~\ref{fig:dem_gap}b. 
Specifically, a pre-trained depth completion network is employed to complement the depth of the 2D image to generate reliable 3D pseudo point clouds. 
\textbf{Second}, we propose a multimodal backbone, Pseudo-Raw Convolution (PRConv), to extract features from pseudo points and raw points while fully enabling the interaction of diverse elements. 
\textbf{Finally}, we propose a Cross-Attention Adaptive Fusion (CAAF), which adaptively fuses multimodal features and constructs correlation information between modalities via a cross-attention variant, which is a more fine-grained fusion strategy.

\section{Proposed Method}
\label{sec:method}

The overall framework of \algorithmname~ is shown in Fig.~\ref{fig:framework}. First, raw points and images yield pseudo points by depth complement, and key points are generated by sampling raw points. Second, the raw points and pseudo points are voxelized. Raw-voxels and pseudo-voxels are fed into PRConv to extract fine-grained features. Final, the heterogeneous RoI features generated by RPN (Region Proposal Network) are fused by CAAF (more detail in Section~\ref{sec:CAAF}) and performing the final regression. 

\subsection{Preliminary}
\label{sec:preliminary}
To convert an image into a pseudo point cloud, pixel-level depth is required. Generally, for a monocular camera, pixel-level depth is obtained by depth estimation or depth complement. Here, we employ the depth complement to obtain more reliable depth information. 
Given a frame of the raw points $P_{raw} \in \mathbb{R} ^ {N \times 3} $ and the RGB image $ I \in \mathbb{R} ^ {W \times H \times 3} $. Here, $N$ is the number of points in the point cloud, and $W$ and $H$ are the width and height of image, respectively. We can project $P_{raw}$ onto image plane with rotation matrix $R \in \mathbb{R}^{3\times3}$ and translation matrix $T\in \mathbb{R}^{3\times4}$ of the LiDAR with respect to the camera reference system to acquire a sparse depth map $S_{depth} \in \mathbb{R}^{W \times H}$ as following: 
\begin{equation}
    S_{depth}(u,v,d) = \mathbf{K} \cdot R \cdot  T \cdot \left [P_{raw}, 1_{N} \right ] ^T
\end{equation}
where $\mathbf{K}$ is the camera intrinsic parameter; $(u,v)$ denote coordinates in pixel coordinate; $d$ is depth value. We define this invertible projection operation as $\phi$. Feeding the $I$ and $S_{depth}$ into a depth complement network $\varphi$ to get a dense depth map $D_{depth} \in \mathbb{R} ^ {W \times H}$ as follow:
\begin{equation}
    D_{depth} = \varphi(I, S_{depth})
    \label{eq:depth_complement}
\end{equation}
Finally, we acquire a frame of pseudo points $P_{pse} \in \mathbb{R} ^ {WH \times 3}$ by inverse projection:
\begin{equation}
    P_{pse} = \phi^{-1} \left ( \left [I, D_{depth} \right ] \right)
\end{equation}
In practice, this process yields decent pseudo points. The definition of the pseudo point will be described in Section~\ref{sec:prconv}.

\subsection{Pseudo-Raw Convolution}
\label{sec:prconv}
\textbf{Pseudo Point} In contrast to the raw point, the pseudo point, which is a more dense representation, contains rich semantic information. Specifically, each pixel of the image will generate a corresponding pseudo point, which consists of the depth ($x, y, z$), the color ($r, g, b$), and the pixel coordinates ($u, v$). 

\textbf{PRConv} Pointnet++~\cite{pointnet++} is a native way for extracting point-wise features. However, due to the vast amounts of pseudo points, it will bring massive calculations by ball query operator~\cite{pointnet++}. To exploit the complementary nature between multimodal points, we propose a point-voxel-based backbone Pseudo-Raw Convolution (PRConv). For the raw point branch, a sparse 3D convolution is used to extract raw-voxel features. Similar to our baseline~\cite{pvrcnn}, the raw-voxel features are converted to 2D BEV (Brid-eye’s view) features by channel transformation and used to obtain proposals. For the pseudo point branch, we initially voxelize it to generate pseudo-voxels and feed them into a 3D sparse convolution to yield the pseudo-voxel features. 

Here, we separately obtain the voxel features of heterogeneous. Although voxel-based features are efficient, inevitable information loss degrades fine-grained localization accuracy. On the other hand, the point-based features allow for fine-grained information but massive calculations. Therefore, we aim to combine the strengths of both methods in an effective way. Inspired by~\cite{pvrcnn}, we perform keypoint sampling by furthest point sample (FPS) in a shared point cloud space to obtain low-noise global keypoints. As shown in Fig.~\ref{fig:framework}, around key points, heterogeneous features are pooled by ball query~\cite{pointnet++} or voxel query~\cite{voxelrcnn} for interaction between features. Finally, $i$-th key point feature $f_{i}^{kp}$ can be represent as following:

\begin{footnotesize}
    \begin{equation}
        f_{i}^{kp}= MLP \left [POOL \left (  f_{i}^{point}, f_{i}^{conv1}, f_{i}^{conv2}, f_{i}^{conv3}, f_{i}^{conv4} \right ) \right ]
    \end{equation}
\end{footnotesize}where $f_{i}^{convk}$ represent $k$-level multimodal semantic features and $f_{i}^{point}$ incorporates both raw point and pseudo point. 
The $MLP(\cdot)$ and $POOL(\cdot)$ means multi-layer perception and max pooling or average pooling, respectively.

\subsection{Cross-Attention Adaptive Fusion}
\label{sec:CAAF}
The nature of LiDAR and cameras leads to an inherent problem---dimension mismatch---which largely limits performance for multimodal detectors. 
Although previous methods~\cite{pointpainting,mvx-net,deepfusion,cat-det} have been proposed to fuse heterogeneous features directly, they fail to address this crucial issue. Benefiting from the generation of pseudo points as mentioned above, the dimension can be nicely matched in a 3D manner. Although the dimensional mismatch can be alleviated by pseudo point, it remains a key challenge regarding alignment. According to our visualization, the pseudo points are distributed with different coordinates and data than the keypoints, which results in their correspondence being complicated rather than aligned one-to-one. 

\begin{figure}
    \centering
    \includegraphics[width=\linewidth]{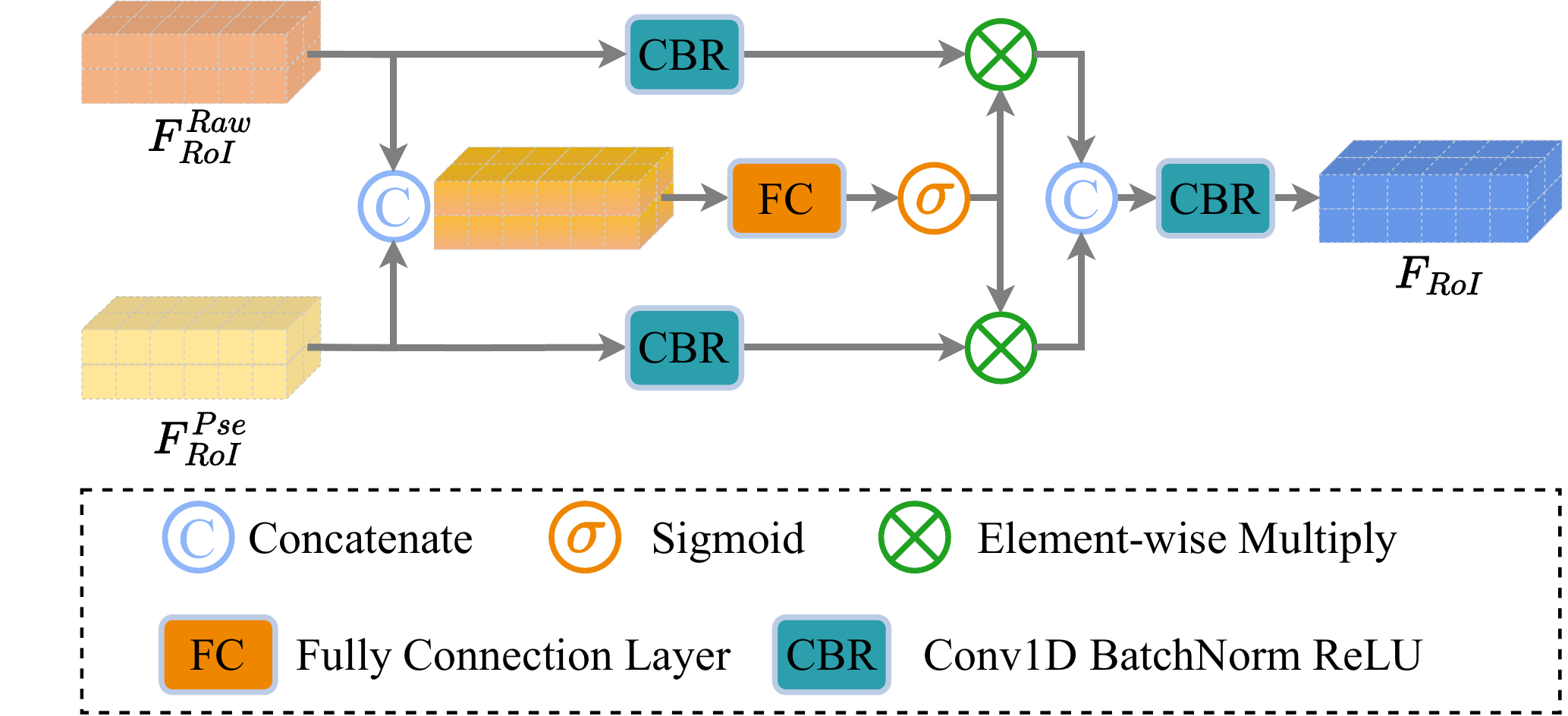}
    \caption{The architecture of CAAF.}
    \label{fig:CAAF}
\end{figure}

To enable the adaptive integration of keypoints and pseudo points, we capture this correspondence dynamically by introducing a cross-attention variant.
Specifically, we utilize CAAF to fuse RoI-wise heterogeneous features in refinement stage. Given a pair of RoIs feature $(F_{RoI}^{raw}, F_{RoI}^{pse}) \in \mathbb{R}^{n \times D_{m}}$, where $n$ is the number of RoIs. We feed the concatenated RoI features into a fully connected layer and a sigmoid function to produce the attention weight $(\mathbf{W}_{raw}, \mathbf{W}_{pse}) \in \mathbb{R} ^ {D_m \times D_m} $. Finally, ($F_{RoI}^{raw}, F_{RoI}^{pse}$) are weighted with $(\mathbf{W}_{raw}, \mathbf{W}_{pse})$ to yield the adaptive fusion of the RoI features $F_{RoI} \in \mathbb{R} ^ {n \times D_m}$. Formally, the CAAF is expressed as following: 
\begin{equation} 
    (\mathbf{W}_{raw}, \mathbf{W}_{pse}) = \sigma(FC(CONCAT(F_{RoI}^{raw}, F_{RoI}^{pse}))) 
\end{equation}
\begin{equation}
    F_{RoI} = CONCAT(\mathbf{W}_{raw} \cdot F_{RoI}^{raw}, \mathbf{W}_{pse} \cdot F_{RoI}^{pse}) 
\end{equation}
where $\sigma(\cdot)$, $FC(\cdot)$, and $CONCAT(\cdot)$ are the operation of the sigmoid function, fully connected layer, and concatenate, respectively. In addition, some dimensional adjustment using convolution 1D.
An illusion is shown in Fig.~\ref{fig:CAAF}.

\subsection{Loss Function}
\label{sec:loss}
Since our framework is a two-stage detector, we setup two fundamental loss, $\mathcal{L}_{RPN}$ and $\mathcal{L}_{Ref}$, for RPN and refinement head, respectively. 
To prevent the gradient from being dominated by a single modal, we designed two auxiliary supervisions, $\mathcal{L}_{AS_{1}}$ and $\mathcal{L}_{AS_{2}}$. 
They are similar with $\mathcal{L}_{Ref}$, containing bounding box regression loss and class confidence loss. The total loss can be represented as following:
\begin{equation}
\mathcal{L}_{Total} = \mathcal{L}_{RPN} + \mathcal{L}_{Ref} + \mathcal{L}_{Depth} + \alpha \mathcal{L}_{AS_{1}} + \beta \mathcal{L}_{AS_{2}}
\end{equation}
where $\alpha$ and $\beta$ weight on $\mathcal{L}_{AS_{1}}$ and $\mathcal{L}_{AS_{2}}$ respectively ($\alpha=0.5$, $\beta=0.5$ by default).
$\mathcal{L}_{Depth}$ is our depth complement network loss.

\begin{table*}[!t]
\footnotesize 
\centering 
\caption{Performance comparison with state-of-the-art methods on nuScenes \textit{test} set without test-time augmentation.  ``C.V.", ``Motor", ``Ped." and ``T.C." means construction vehicle, motorcycle, pedestrian and traffic cone, respectively. Best in bold.} 
\label{tab:nuscenes}
\begin{tabular}{|c|cccccccccc|cc|}
\hline
Method & Car & Truck & C.V. & Bus & Trailer & Barrier & Motor. & Bike & Ped. & T.C. & mAP & NDS \\ \hline\hline
CBGS~\cite{cbgs}  & 81.1  & 48.5  & 10.5 & 54.9  & 42.9 & 65.7 & 51.5 & 22.3 & 80.1 & 70.9 & 52.8 & 63.3 \\
CenterPoint~\cite{centerpoint}  & 84.6 & 51 & 17.5 & 60.2 & 53.2 & 70.9 & 53.7 & 28.7 & 83.4 & 76.7 & 58.0 & 65.5 \\
VoxelNeXt~\cite{voxelnext}  & 84.6 & 53.0 & 28.8 & 64.7 & 55.8 & 74.6  & \textbf{73.2} & 45.7  & 85.9 & 79.0 & 64.5  & 70.0\\
LargeKernel3D~\cite{largekernel3d} & 85.9 & 55.3  & 26.8  & 66.2 & \textbf{60.2} & 74.3 & 72.5 & 46.6 & 85.6  & 80.0 & 65.3 & 70.5 \\
VISTA~\cite{vista} & 84.7 & 54.2  & \textbf{29.1} & 64.0 & 55.0    & 71.8 & 71.0 & 45.2  & 83.6 & 78.6 & 63.7 & 70.4 \\ 
MVP~\cite{mvp} & \textbf{86.8} & 58.5 & 26.1 & 67.4 & 57.3 & 74.8 & 70.0 & 49.3 & 89.1 & 85.0 & 66.4 & 70.5 \\ \hline\hline
\textbf{Ours} & \textbf{86.8} & \textbf{58.7} & 26.6 & \textbf{68.0} & 57.4 & \textbf{74.9} & 70.8 & \textbf{54.0} & \textbf{89.4} & \textbf{85.6} & \textbf{67.2}  & \textbf{71.0} \\ \hline
\end{tabular}
\end{table*}

\begin{table}[!t]
    \footnotesize 
    \centering 
    \caption{Performance comparison with state-of-the-art methods on KITTI \textit{val} set for car category. ``Mod." and ``-" means moderate and not mention, respectively. Best in bold.} 
    \renewcommand\arraystretch{1.5} 
    \setlength{\tabcolsep}{0.9mm}{ 
    \begin{tabular}{|c|c|cccccc|} 
    \hline 
    \multirow{2}{*}{Methods} & 
    \multirow{2}{*}{Input} &
    \multicolumn{3}{c}{AP$_{3D}$(\%)} &
    \multicolumn{3}{c|}{AP$_{BEV}$(\%)} \\ \cline{3-8} 
    & & Easy & Mod. & Hard & Easy & Mod. & Hard \\ \hline \hline 
    PV-RCNN~\cite{pvrcnn} & - & 92.57 & 84.83 & 82.69 & 95.76 & 91.11 & 88.93 \\ 
    Voxel R-CNN~\cite{voxelrcnn} & - & 92.38 & 85.29 & 82.86 & 95.52 & 91.25 & 88.99 \\ 
    EPNet~\cite{epnet} & 3D+2D & 92.28 & 82.59 & 80.14 & 95.51 & \textbf{91.47} & \textbf{91.16} \\ 
    CAT-DET~\cite{cat-det} & 3D+2D & 90.12 & 81.46 & 79.15 & - & - & - \\
    Focals Conv-PV~\cite{focalsconv} & 3D+2D & 92.26 & 85.32 & 82.95 & - & - & - \\
    EQ-PVRCNN~\cite{eqpvrcnn} & 3D+2D & 92.52 & 85.61 & 83.13 & - & - & - \\ \hline \hline
    \textbf{Ours} & 3D+3D & \textbf{95.26} & \textbf{85.84} & \textbf{83.67} & \textbf{95.91} & 89.65 & 89.43 \\ \hline
    \end{tabular}}
    \label{tab:kitti_val}
\end{table}

\section{Experiments} 
\subsection{Dataset and Implementation Details} 
We conduct experiments on two popular autonomous driving datasets, i.e., KITTI~\cite{KITTI} and nuScenes~\cite{nuscenes}.
The KITTI contains 7481 annotated samples and we split it into \textit{train} set with 3712 samples and \textit{val} set with 3769 samples following popular 3D detection models~\cite{pvrcnn}. As to the nuScenes, a large-scale autonomous driving dataset, it contains 700 training scenes, 150 validation scenes and 150 test scenes. Each scene has approximately 200 frames. 
For KITTI, we evaluate the result on \textit{val} set with average precision (AP) calculated by 40 recall positions on easy, moderate and hard difficulties~\cite{KITTI}. For nuScenes, we evaluate the performance of different models using the official metrics, i.e., mean Average Precision (mAP) and nuScenes detection score (NDS)~\cite{nuscenes}.

We leverage PENet~\cite{penet} and MVP~\cite{mvp} to generate pseudo points for KITTI and nuScenes, respectively. Though the depth complement can be processed in an end-to-end manner, we adopt an offline way of processing with pseudo point clouds, which allows us to conduct experiments rapidly.  
For PRConv, we set the number of keypoints to 2048, 
and 4096 for kitti and nuScenes, respectively. We implement our method using the open-source framework OpenPCDet~\cite{openpcdet}. All the experiments trained on 8 RTX 3090 GPUs and inferred on a single RTX 3090 GPU. 
To ensure fairness, we follow the baseline, PV-RCNN~\cite{pvrcnn}, to set hyperparameters as specified in OpenPCDet~\cite{openpcdet}. 
Specifically, \algorithmname~uses the Adam optimizer with a learning rate of $1 \times 10^{- 2}$ and adopts the one-cycle learning scheme. 

\subsection{Main Results} 
To verify the effectiveness of our method, we compare it with several state-of-the-art methods on nuScenes~\cite{nuscenes}. As shown in Table~\ref{tab:nuscenes}, ~\algorithmname~outperforms all previous state-of-the-arts on most metrics. Especially, it outperforms MVP~\cite{mvp} by 0.8\% mAP and 0.5\% NDS on \textit{test} set, respectively. In addition, we note that it outperforms MVP in all categories and has a significant improvement over the difficult categories (e.g., C.V., Bike). 
As shown in Table~\ref{tab:kitti_val}, \algorithmname~improves baseline~\cite{pvrcnn} by 2.69\% $AP_{3D}$ in easy level benefiting from 3D+3D fusion manner, which further demonstrates the effectiveness of our approach on KITTI~\cite{KITTI}. 

\begin{table}[!t]
    \footnotesize
    \centering
    \caption{Ablation Experiments of Different Module on KITTI \textit{val} set. PRConv and CAAF stand for Pseudo-Raw Convolution and Cross-Attention Adaptive Fusion respectively.}
    \renewcommand\arraystretch{1.5}
    \setlength{\tabcolsep}{0.9mm}{
    \begin{tabular}{|c|cc|cccccc|}
    \hline
    \multirow{2}{*}{Experiment} &
    \multirow{2}{*}{PRConv} &
    \multirow{2}{*}{CAAF} &
    \multicolumn{3}{c}{AP$_{3D}$(\%)} &
    \multicolumn{3}{c|}{AP$_{BEV}$(\%)} \\ \cline{4-9} 
    & & & Easy & Mod. & Hard & Easy & Mod. & Hard \\ \hline \hline 
     (a) & & & 91.84 & 82.93 & 82.24 & 92.88 & 90.39 & 88.39 \\ 
    (b) & \checkmark & & 92.87 & 83.62 & 82.85 & 95.87 & 89.38 & 88.98 \\  
    (c) & \checkmark & \checkmark & \textbf{95.26} & \textbf{85.84} & \textbf{83.67} & \textbf{95.91} & \textbf{89.65} & \textbf{89.43}\\ \hline 
    \end{tabular}} 
    \label{tab:ablation} 
\end{table}

\begin{table}[!t]
    \scriptsize 
    \centering
    \caption{Effects of different feature components for PRConv module. ``Mod." means moderate.}
    \renewcommand\arraystretch{1.5}
    \setlength{\tabcolsep}{0.9mm}{
    \begin{tabular}{|c|ccccc|ccc|}
    \hline
    \multirow{2}{*}{Experiment} &
          \multirow{2}{*}{$f_{i}^{point}$} 
          & \multirow{2}{*}{$f_{i}^{conv1}$} 
          & \multirow{2}{*}{$f_{i}^{conv2}$} & \multirow{2}{*}{$f_{i}^{conv3}$} & \multirow{2}{*}{$f_{i}^{conv4}$} & \multicolumn{3}{c|}{AP$_{3D}$(\%)} \\ \cline{7-9}
          & & & & & & Easy & Mod. & Hard \\ \hline \hline
        (a) & \checkmark & & & & & 92.44 & 83.22 & 80.92 \\  
        (b) &
         \checkmark & \checkmark & & & & 92.92 & 83.72 & 81.24 \\
         (c) &
         \checkmark & \checkmark & \checkmark & & & 92.92 & 84.02 & 83.41 \\
         (d) &
         \checkmark & \checkmark & \checkmark & \checkmark & & 93.13 & 85.72 & 83.45 \\
         (e) &
         \checkmark & \checkmark & \checkmark & \checkmark & \checkmark & \textbf{95.26} & \textbf{85.84} & \textbf{83.67}  \\ \hline
    \end{tabular}}
    \label{tab:ablation2}
\end{table}

\subsection{Ablation Studies and Analysis} 
\label{sec:ablation} 
\textbf{Component-Wise Analysis.} 
To verify the efficiency of each component, we incrementally conduct component-wise experiments on the baseline~\cite{pvrcnn}. As shown in Tab.~\ref{tab:ablation}, in experiment (a), we apply the pseudo point solely to the baseline with only a slight performance improvement. 
In experiment (b), PRConv extract pseudo point features and exploit interaction between two modalities, making 1.03\%, 0.69\% and 0.61\% enhancment on easy, moderate and hard respectively. Experiment (c) equip CAAF module on experiment (b) that fusion isomerous 3D RoI features adaptively, achieving best performance on all mercies. It is notable that easy and moderate in $AP_{3D}$ received a significant improvement of 2.39\% and 2.22\% respectively, which demonstrates the efficiency of CAAF.

\textbf{Effects of different features for PRConv.} 
In Table~\ref{tab:ablation2}, we explore the effect of aggregating different multimodal features on the PRConv. In experiment (a), it shows that the performance drops considerably if only aggregating $f_{mm}^{point}$, due to the shallow level of modal interaction not being sufficient to support higher performance. In experiments (b), (c), (d) and (e), we incrementally apply high-level semantic features until reaching peak performance. It shows that high-level semantic features can make a big difference in performance and proves that PRConv extracts informative multimodal features.

\section{Conclusion} 
In this paper, we proposed a multimodal 3D object detection framework~\algorithmname, which leverages pseudo point generated by depth complement to tackle dimension mismatch caused by point clouds and images. With the proposed PRConv, the multimodal features are efficiently and adequately extracted as multimodal interactions arise. Besides, we design a CAAF module that adaptively integrates multimodal features through a cross-attention mechanism. 
Experiments are conducted on KITTI and nuScenes, demonstrating our method improves accuracy prominently. 

\newpage

\bibliographystyle{IEEEtran}
\bibliography{FGU3R}

\end{document}